\documentclass{article}

\usepackage{PRIMEarxiv}
\usepackage[utf8]{inputenc}
\usepackage[T1]{fontenc}
\usepackage{textcomp}
\usepackage{graphicx}
\usepackage[table]{xcolor}
\usepackage{hyperref}
\usepackage{url}
\usepackage{booktabs}
\usepackage{array}
\usepackage{amsmath}
\usepackage{amssymb}
\usepackage{nicefrac}
\usepackage{caption}
\usepackage{microtype}
\usepackage{placeins}

\graphicspath{{media/}}
\definecolor{linkblue}{rgb}{0,0,0.45}
\hypersetup{colorlinks=true,citecolor=linkblue,linkcolor=linkblue,urlcolor=linkblue}

\definecolor{gtmNavy}{HTML}{002456}
\definecolor{gtmBeige}{HTML}{E1C1A4}
\definecolor{gtmLightBlue}{HTML}{64B9EE}
\definecolor{gtmLightGreen}{HTML}{9BD18B}
\definecolor{gtmLightCoral}{HTML}{F08080}
\definecolor{gtmLightPurple}{HTML}{B8A3D9}
\newcommand{\methodcolor}[2]{\raisebox{0.15ex}{\textcolor{#1}{\rule{0.9em}{0.9em}}}~#2}

\title{Acquisition Geometry-Assisted Whole-Group Localization of X-ray Fluorescence Maps in Optical Microscopy Images}

\author{
  Xiangyu Yin$^{1,*}$, Tatjana Paunesku$^{2}$, Letonia Copeland-Hardin$^{2,3}$,
  Martina Ralle$^{4}$, \\[0.25em]
  \textbf{Zichao Wendy Di$^{1}$, Si Chen$^{1}$, Gayle E. Woloschak$^{2}$,
  Barry Lai$^{1}$, Mathew J. Cherukara$^{1}$, and Stefan Vogt$^{1,\dagger}$} \\[0.6em]
  $^1$Advanced Photon Source, Argonne National Laboratory, Lemont, Illinois, USA \\
  $^2$Northwestern University, Chicago, Illinois, USA \\
  $^3$University of Chicago, Chicago, Illinois, USA \\
  $^4$Oregon Health and Science University, Portland, Oregon, USA \\
  $^*$\texttt{xyin@anl.gov}; $^\dagger$\texttt{svogt@anl.gov}
}

\begin{document}
\pagenumbering{gobble}
\thispagestyle{empty}
\textbf{GOVERNMENT LICENSE}

The submitted manuscript has been created by UChicago Argonne, LLC, Operator of Argonne National Laboratory (“Argonne”). Argonne, a U.S. Department of Energy Office of Science laboratory, is operated under Contract No. DE-AC02-06CH11357. The U.S. Government retains for itself, and others acting on its behalf, a paid-up nonexclusive, irrevocable worldwide license in said article to reproduce, prepare derivative works, distribute copies to the public, and perform publicly and display publicly, by or on behalf of the Government. The Department of Energy will provide public access to these results of federally sponsored research in accordance with the DOE Public Access Plan. \href{http://energy.gov/downloads/doe-public-access-plan}{http://energy.gov/downloads/doe-public-access-plan}
\clearpage
\pagenumbering{arabic}

\maketitle

\begin{abstract}
X-ray fluorescence (XRF) microscopy maps elemental distributions, while optical microscopy can provide complementary morphological context. Localizing XRF fields of view (FOVs) in optical images is difficult because the two modalities differ in contrast mechanism and resolution. Most current workflows place each XRF tile independently, even when acquisition metadata already record the tiles' relative scan positions. This study formalizes XRF tile-group localization, in which one optical-frame placement is estimated for the whole group, constrained by acquisition geometry and quantified using group intersection-over-union (GroupIoU). In a controlled case study, independent localization failed with GroupIoU 0.000, whereas group localization achieved 0.931. Replacing the normalized cross-correlation (NCC) metric with mutual information (MI) gave nearly identical results, showing that the outcome is not specific to one local similarity metric. In another multiscale case study, using a coarse XRF survey scan to connect the fine-scale tile group to the optical image increased mean GroupIoU from 0.694 to 0.856. These case studies support using acquisition geometry as an explicit constraint when localizing related XRF tiles.
\end{abstract}


\section{Introduction}
\label{sec:introduction}

X-ray fluorescence (XRF) microscopy reveals the spatial distribution of elements in biological, materials science, and environmental specimens at sub-micrometer resolution~\cite{pushie2014xrf,fahrni2007biological,dejonge2010xrf}. It can measure trace metal concentrations, drug accumulation patterns, and contaminant localization, information that is difficult to obtain by other means, which makes XRF a widely used tool at synchrotron light sources and in laboratory instruments~\cite{copeland2023proof,yuan2013epidermal,webb2022xrf,strotton2023ztag}. The value of these elemental maps grows substantially when they can be placed in morphological context: which cell type accumulates the metal? which tissue layer contains the contaminant? Answering such questions requires correlating XRF maps with optical microscopy images that provide the structural information missing from the XRF maps~\cite{caplan2011correlative,mcrae2006microxrf,ftir_xrf_2022}.

Localizing XRF maps within optical images, however, remains challenging~\cite{sotiras2013deformable,song2017medical}. The two modalities differ in resolution, FOV, and most fundamentally in the contrast mechanism: optical microscopy records visible-light absorption and scattering, while XRF records characteristic X-ray emission from specific elements. Standard intensity-based methods often fail if applied directly, and feature descriptors designed for single-modality images rarely transfer across this contrast gap. Mutual information~\cite{maes1997mutual,pluim2003survey}, fiducial markers~\cite{sheriff2021autocrim}, landmark-based tools~\cite{paulgilloteaux2017ecclem}, and learned representations~\cite{haskins2020deep,cao2014analogies} each address some aspects of this challenge, but common localization workflows reduce the problem to a single image pair or a single tile.

We argue that this reduction discards valuable acquisition-geometry information that the instrument already records. At synchrotron beamlines, high-resolution elemental maps are acquired as raster scans over defined regions of interest. When multiple regions are scanned on the same specimen, the instrument records their relative positions as standard acquisition metadata. Multiscale workflows contribute additional structure: a coarse survey scan may cover a large area, with finer target scans nested inside it~\cite{xct_xrf_2024,albers2021elastic,mei2025bigreg}. The result is a geometrically constrained group of tiles instead of a collection of independent tiles.

The idea explored in this paper is that the group, not the individual tile, is the natural unit of localization. We first formalize this idea and formulate geometry-aware group localization, including two group-localization strategies and multiscale variants that add a bridge image. We then develop a geometry-aware matcher based on these ideas and test it on two case studies. The first is an isolation case study designed so that all variables other than acquisition geometry are held fixed, so that any difference can be attributed to geometry. The second is a multiscale case study on four specimen groups under more typical experimental conditions, in which an intermediate coarse XRF scan acts as a bridge between the target tile group and the optical frame.

\section{Related Work}
\label{sec:related_work}

\paragraph{Correlative microscopy workflows.}
Correlative microscopy localization has been developed most actively for correlative light and electron microscopy (CLEM), producing tools such as eC-CLEM for interactive landmark alignment~\cite{paulgilloteaux2017ecclem} and autoCRIM for fiducial-based automation~\cite{sheriff2021autocrim}. Cross-modal alignment remains a common bottleneck across many modality combinations~\cite{walter2020correlative}. In correlative X-ray microscopy, workflows combining XRF with X-ray tomography~\cite{xct_xrf_2024}, infrared spectroscopy~\cite{ftir_xrf_2022}, or histology~\cite{albers2021elastic} often depend on manual overlays or case-specific alignment steps. Recent automated pipelines~\cite{mei2025bigreg} improve throughput for paired images, and dedicated XRF-to-RGB localization methods have been developed for single-pair settings~\cite{amiri2025xrf,bock2023registration}. ML-assisted scanning frameworks such as ROI-Finder use learned models to guide XRF acquisition itself~\cite{chowdhury2022roi}. The gap addressed here is different: in XRF-to-optical experiments, the target is often a group of XRF tiles with known relative positions rather than a single independent image pair.

\paragraph{Image-based multimodal registration.}
Cross-modal registration is a long-standing challenge in biomedical imaging~\cite{sotiras2013deformable}. Intensity-based methods compare image statistics, with mutual information a widely used search objective~\cite{maes1997mutual,pluim2003survey}. Feature-based methods match local descriptors, but descriptors can fail when two modalities show different physical structures~\cite{song2017medical}. Learned methods estimate transformations or shared representations from training data~\cite{haskins2020deep,cao2014analogies}, with recent work using unsupervised deep registration~\cite{grexa2024supercut} and contrastive multimodal representations for cross-modal sub-image retrieval~\cite{breznik2024crossmodality}. These methods draw their evidence mainly from image appearance rather than from landmarks or fiducials. Our work also depends on image appearance-based evidence but adds a separate source of information: the instrument-recorded geometry that links the XRF tiles before localization begins.

\paragraph{Geometry as localization information.}
Known spatial relationships are already used in other localization settings. Microscopy stitching uses stage positions to constrain translations between same-modality image tiles~\cite{preibisch2009stitching,chalfoun2017mist}. Atlas-based neuroimaging uses anatomical shape models~\cite{goubran2019connectomic}. Panoramic photography and remote sensing use camera geometry to place multiple images jointly. These examples show that geometry can turn several separate localization problems into one constrained placement problem. XRF acquisition metadata provide this type of constraint, but prior XRF-to-optical localization workflows have not treated the XRF tile group itself as the localization object. Stitching uses geometry within one modality to compose images, while we use geometry across modalities to localize a group.

\section{Methods}
\label{sec:methods}

In this work, we treat localization as the placement of an entire XRF tile group in the optical coordinate frame. This section formulates the localization task and its independent per-tile reference (\S\ref{sec:problem_formulation}), describes the group-localization score and the two single-scale strategies it supports (\S\ref{sec:group_localization_strategies}), and extends the formulation to a multiscale setting with an intermediate coarse XRF scan (\S\ref{sec:multiscale_localization}).

\subsection{Problem Formulation}
\label{sec:problem_formulation}

\begin{figure}[!htbp]
  \centering
  \includegraphics[width=0.85\linewidth]{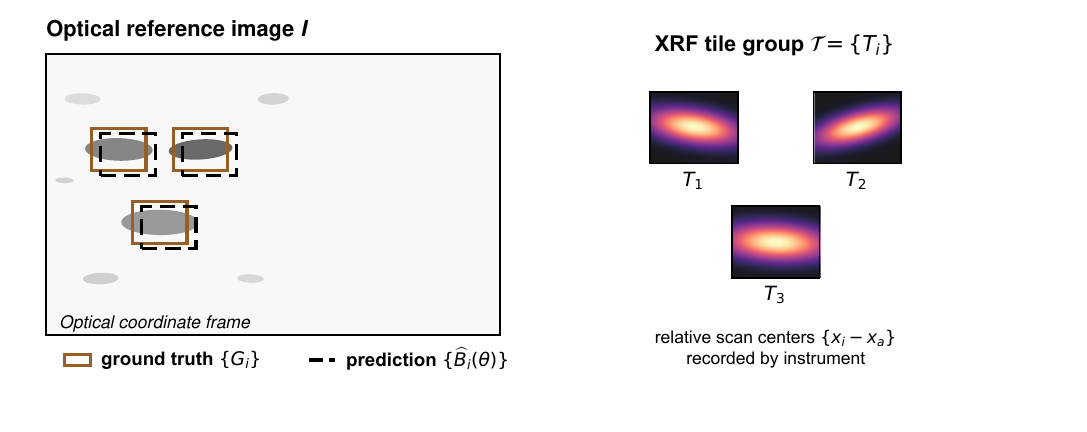}
  \caption{Task overview. The XRF acquisition produces a group of high-resolution tiles whose relative positions are recorded by the instrument. The goal is to estimate one placement that maps the tile group into the optical coordinate frame so that all tile footprints are jointly located.}
  \label{fig:task_overview}
\end{figure}

Figure~\ref{fig:task_overview} summarizes the localization task. The acquisition metadata specify the relative tile positions, while localization estimates one placement that maps the tile group into the optical coordinate frame.

Let $I$ denote the optical reference image and $\mathcal{T}=\{T_1,\dots,T_n\}$ an XRF tile group acquired from the same specimen. The instrument records, for each tile $T_i$, a physical scan center $x_i\in\mathbb{R}^2$ and the per-pixel size of that scan. The localization goal is to estimate a group placement $\theta$ that maps every tile $T_i$ into the optical coordinate frame at the same time. We use \emph{footprint} to refer to the rectangular region a tile occupies in optical pixels under a candidate placement (notation collected in Appendix~\ref{app:math_formulation}, Table~\ref{tab:method_notation}).

A placement $\theta$ is specified by five components: the \emph{anchor} (index $a$), the anchor center $c_a(\theta)$ in optical pixels, a global scale, an orientation, and an optional flip. From the recorded scan metadata we compute an \emph{offset map} $P_\theta$ that converts a physical offset between two scan centers into the corresponding offset in optical pixels. The predicted center of any other tile $T_i$ under that placement is
\begin{equation}
\widehat{c}_i(\theta)=c_a(\theta)+P_\theta\left(x_i-x_a\right),
\label{eq:center_prediction}
\end{equation}
The predicted footprint $\widehat{B}_i(\theta)$ is the rectangle centered at $\widehat{c}_i(\theta)$ with size set by the candidate scale and the recorded tile pixel sizes. Localization then selects the placement $\theta$ that best scores the whole tile group under a given score function $S_{\mathrm{grp}}$ (defined in Section~\ref{sec:group_localization_strategies}) .

We assume that all tiles share one global similarity transform comprising translation, rotation, and uniform scaling, with an optional reflection. This transform preserves the relative layout $\{x_i-x_a\}$ recorded by the instrument. Nonrigid deformations are outside the scope of the present study.

The conventional approach treats XRF tiles as independent localization targets rather than as a constrained group. We use this conventional independent per-tile localization as the non-geometric reference baseline: each tile is matched separately, without acquisition-geometry coupling during scoring, and the predicted footprints are unioned only for reporting.

\subsection{Group localization and single-scale strategies}
\label{sec:group_localization_strategies}

We seek a score for a candidate placement $\theta$ that rises when all tiles in $\mathcal{T}$ can be placed plausibly at once, and fails when the placement is inconsistent with the recorded acquisition geometry. Given $\theta$, we consider each tile's predicted footprint $\widehat{B}_i(\theta)$, search a small window around it, and accept the best local match $B_i^\star(\theta)$ as the tile's locally verified footprint. The window is deliberately narrow: each companion tile is given enough freedom to absorb minor calibration error but cannot drift beyond the position implied by the recorded geometry. The distance from $B_i^\star(\theta)$ back to $\widehat{B}_i(\theta)$ is the \emph{center residual} $r_i(\theta)$.

A tile only contributes if it passes basic checks (its local image score is high enough, its footprint lies inside $I$, and its weight is above a minimum). The whole-group score $S_{\mathrm{grp}}(\theta)$ then combines three ingredients:
\begin{enumerate}
  \item the average local image similarity over verified tiles, normalized by the expected tile weight so that partial matches accumulate less score than full matches,
  \item a penalty for tiles whose verified position drifted far from the recorded prediction (smaller residuals are better),
  \item a penalty for disagreement between the predicted and verified tile-overlap patterns.
\end{enumerate}
Candidate placements that fail on tile count, matched weight, or visibility are returned as a no-match rather than a low-confidence guess. The full functional form, the weights, and the normalizers are given in Appendix~\ref{app:math_formulation}. 

Two strategies instantiate this idea. They differ in \emph{when} the acquisition geometry enters the computation.

\paragraph{Mosaic localization.}
The tiles are first fused into a single template in their group coordinate frame, using the recorded geometry to place them relative to one another. The fused mosaic is then matched against the optical image as if it were a single template. Acquisition geometry is used during template construction, but once the mosaic exists, individual tiles cannot shift to absorb residual calibration error. This is the ``hard-constraint'' end of the geometry prior spectrum.

\paragraph{Anchor-verified group localization.}
The tiles are kept separate throughout scoring. A small set of candidate anchor placements is generated from the anchor tile's appearance (by template localization followed by non-maximum suppression), and each candidate is scored by $S_{\mathrm{grp}}$ over the full group. Because $S_{\mathrm{grp}}$ uses the recorded geometry as a soft constraint inside the local-verification window, individual tiles can absorb small calibration errors while the group as a whole is still required to be consistent. This is the ``soft-constraint'' end of the geometry prior spectrum.

\subsection{Multiscale group localization}
\label{sec:multiscale_localization}

Many XRF experiments record an intermediate coarse survey scan $J$ that covers a larger region than the target tile group but a smaller region than the optical image. We treat $J$ as a \emph{bridge}: an additional level in the group that connects the target tile group to the optical frame. Two multiscale strategies use the bridge differently. They are parallel members of the group-localization family above, just with one more scale.

\paragraph{Structural-bridge localization.}
The bridge is first matched to the optical frame to estimate where the bridge sits in optical coordinates. The target group is then matched inside the bridge image (reusing the same group score $S_{\mathrm{grp}}$ with the bridge $J$ standing in for the optical reference), and the two placements are composed to land the target group in the optical frame. At a high level, this route estimates the bridge-to-optical placement $\widehat{\psi}$ and the target-to-bridge placement $\widehat{\omega}$, then returns the composed optical-frame placement $\widehat{\theta}_{\mathrm{struct}}=\widehat{\psi}\circ\widehat{\omega}$. This route is useful when the bridge is easier to localize in the optical frame than the small target tiles are, but it depends on the bridge-to-optical step succeeding.

\paragraph{Bridge-prior direct localization.}
The bridge is used as a spatial prior instead of a coordinate frame. The bridge-to-optical estimate, combined with the recorded bridge-to-target metadata, implies an approximate optical-frame location $\theta_0$ for the target group. We then run the group matcher directly against the optical image but restrict the search to a neighborhood around $\theta_0$. At a high level, this route returns $\widehat{\theta}_{\mathrm{prior}}$ by maximizing $S_{\mathrm{grp}}$ within that neighborhood, with an optional penalty for distance from $\theta_0$. The final localization stays in the optical frame, but the bridge narrows the search to a sensible region. This route is useful when the bridge gives a good approximate location yet its appearance differs enough from $I$ that we do not want it to drive the final placement.

Because the two strategies are complementary, we run both and select between them with a runtime-only rule:
\begin{equation}
\widehat{\theta}_{\mathrm{bridge}}=
\begin{cases}
\widehat{\theta}_{\mathrm{struct}}, & \text{if the structural-bridge route returns a runtime-valid group result},\\
\widehat{\theta}_{\mathrm{prior}}, & \text{otherwise}.
\end{cases}
\label{eq:bridge_rule}
\end{equation}
A route is ``runtime-valid'' only if it returns a placement that passes computational checks on matched tile count, matched tile weight, and geometric visibility. A predicted tile is counted as visible when at least 5\% of its footprint lies within the reference-image bounds, and the multiscale experiments require at least 60\% of the tile count and tile weight to be visible. These checks do not involve human visual judgment, ground-truth overlap, group identity, or tile-level IoU. If a route cannot produce a placement that passes the checks, it returns no match. The full route objectives are given in Eqs.~\ref{eq:structural_bridge} and~\ref{eq:bridge_prior_direct}.

\section{Experiment Design}
\label{sec:experiment_design}

\subsection{Datasets for case studies}
\label{sec:datasets}

We evaluate on two real datasets acquired at the Advanced Photon Source (APS). The controlled benchmark comprises a single optical reference image together with several XRF scans of ROIs within the same specimen, each fit with MAPS to produce 26 elemental channels on an 81$\times$81 or 91$\times$91 scan grid at a 10\,\textmu{}m step size. The isolation case study (Section~\ref{sec:isolation_result}) uses a two-tile XRF scan group drawn from this benchmark. The same two-tile pair is also the basis for a designed perturbation protocol (Section~\ref{sec:perturbation_protocols}) that characterizes how each strategy responds to specific failure modes. 

The multiscale dataset comprises four specimen groups (Case 1--4), each with an optical image, a coarse XRF bridge image, and several higher-resolution, smaller-FOV XRF tiles. This dataset is discussed in Section~\ref{sec:multiscale_result}. For this dataset, the optical image and the corresponding XRF scans show adjacent serial sections from the same specimen rather than the same physical section. The sample preparation and imaging procedures closely resemble those described in \cite{copeland2023proof}. In brief, archival formalin-fixed, paraffin-embedded (FFPE) lymph node and lung tissues from beagle dogs were retrieved using archival information from the Northwestern University Radiobiology Archive (NURA\footnote{https://sites.northwestern.edu/nura/data/inhalation-toxicology-research-institute-data/}). Two immediately adjacent serial tissue sections, each 5--10~\textmu{}m thick, were generated using a microtome. One section was placed on a glass slide, stained with hematoxylin and eosin (H\&E), and imaged with a Hamamatsu NanoZoomer single-slide imager. The adjacent section was placed on an Ultralene membrane (SPEX) and imaged by XRF microscopy. Low-resolution survey scans were acquired at beamline 8-BM-B using 11.2~keV X-rays focused to a 30~\textmu{}m spot, with a dwell time of 50~ms per pixel. An SII Vortex ME4 four-element silicon drift detector collected the spectra. After imaging the full sample area, the samples were physically trimmed in two dimensions, and the smaller specimens were imaged at the medium-resolution beamline 2-ID-E. An X-ray energy of 10.5~keV was used, and the beam was focused to a 0.5~\textmu{}m spot using Fresnel zone plate optics. Raster scans used a 1~\textmu{}m step size and a 50~ms dwell time per pixel. Experts manually determined the ground-truth locations from tissue and empty-space patterns together with the sample-trimming information, without fiducial markers.

\subsection{Image representations and search space}
\label{sec:representations_search_space}

All methods operate on normalized 2D grayscale representations, so that optical contrast and XRF elemental contrast share a single comparable channel for cross-modal scoring and the effect of acquisition geometry is not entangled with the choice of image features. Optical images are read from TIFF and converted to grayscale at load time. XRF maps are read from HDF5 files, percentile-normalized (2--98\%), and resampled to an isotropic grid. We use gradient magnitude for cross-modal steps and raw XRF intensity for within-modality steps. Because the XRF map carries multiple elemental channels, we run the isolation case study under two channel-selection settings: \emph{auto-channel} (top-$k$ channels by intensity variance, default $k{=}5$, combined into a single grayscale template) and \emph{fixed-P} (phosphorus channel only, broadly present in tissues).

The search space spans scale, orientation, and translation. Scale is centered on instrument calibration metadata rather than on ground truth, and orientation is searched coarse-to-fine around the nominal angle. Candidate placements are scored by local image similarity and then filtered by group-level constraints. Full parameter settings are documented in Appendix~\ref{app:case_level_diagnostics}.

\subsection{Baseline and controls}
\label{sec:baselines_controls}

The non-geometric reference is the independent per-tile localizer defined in \S\ref{sec:problem_formulation}, which represents the conventional approach of registering XRF tiles as independent templates. The two group-localization strategies (mosaic and anchor-verified) and the two multiscale strategies (structural bridge and bridge-prior direct) are the methods under study. Table~\ref{tab:strategy_definitions} summarizes the strategy variants, their figure colors, whether acquisition geometry enters the score, and how the bridge image is used.

\begin{table}[!htbp]
  \centering
  \caption{Method variants used in this study. The color column defines the method palette used in the result figures.}
  \label{tab:strategy_definitions}
  \small
  \setlength{\tabcolsep}{2.3pt}
  \begin{tabular}{>{\raggedright\arraybackslash}p{0.17\linewidth} >{\raggedright\arraybackslash}p{0.14\linewidth} >{\raggedright\arraybackslash}p{0.25\linewidth} >{\raggedright\arraybackslash}p{0.16\linewidth}  >{\raggedright\arraybackslash}p{0.16\linewidth}}
    \toprule
    Strategy & Color & Candidate source & Geometry in score & Bridge use \\
    \midrule
    Single-tile reference & \methodcolor{gtmNavy}{navy} & Independent per-tile appearance & No & None \\
    Mosaic & \methodcolor{gtmBeige}{beige} & Fused group template & During template construction & None \\
    Anchor-verified group & \methodcolor{gtmLightBlue}{blue} & Anchor candidates & Yes & None \\
    Structural bridge & \methodcolor{gtmLightGreen}{green} & Bridge-to-optical plus group-in-bridge & Yes & Bridge reference frame \\
    Bridge-prior direct & \methodcolor{gtmLightCoral}{coral} & Bridge-seeded optical-frame search & Yes & Bridge spatial prior \\
    Bridge rule & \methodcolor{gtmLightPurple}{purple} & Structural if runtime-valid, otherwise bridge-prior direct & Yes & Runtime route selection \\
    \bottomrule
  \end{tabular}
\end{table}

Two controls help isolate why the group methods help. On the isolation case study, we replace normalized cross-correlation (NCC) with mutual information (MI)~\cite{maes1997mutual,pluim2003survey} in $\phi_i$ to test whether the result depends on a particular local similarity criterion. On the multiscale case study, we evaluate off-the-shelf SIFT and ORB feature localization~\cite{lowe2004sift,rublee2011orb} to test whether a standard keypoint pipeline could replace the template-and-geometry combination. For each control, we hold the rest of the search procedure fixed and change one source of information at a time. Settings are documented in Appendix~\ref{app:case_level_diagnostics}.

\subsection{Metrics}
\label{sec:metrics}

The primary metric is GroupIoU, the intersection-over-union of the predicted and ground-truth tile unions:
\begin{equation}
\mathrm{GroupIoU}=\frac{\left|\left(\bigcup_i B_i\right)\cap\left(\bigcup_i G_i\right)\right|}{\left|\left(\bigcup_i B_i\right)\cup\left(\bigcup_i G_i\right)\right|}.
\label{eq:group_iou}
\end{equation}
Here $B_i$ is the predicted footprint for tile $T_i$ and $G_i$ is its ground-truth footprint. For independent per-tile localization, each tile is matched separately to produce its own $B_i$, and the same union formula is then applied. 

We also report TileIoU, the ordinary IoU computed separately for each tile and summarized by minimum, mean, and weighted mean. This catches failures that GroupIoU can hide. For example, if a two-tile predicted union covers most of the correct field of view but one individual tile is assigned to the wrong part of that union, GroupIoU can still be moderate or high while the minimum TileIoU is 0. We therefore use GroupIoU for whole-FOV placement and TileIoU summaries to expose which part of the group succeeded or failed.

\subsection{Perturbation protocols on the isolation case study}
\label{sec:perturbation_protocols}

To characterize how each strategy responds to specific failure modes, we apply two perturbation protocols on the controlled two-tile pair. The metadata-position jitter sweep draws Gaussian perturbations of the companion tile's recorded scan center across 21 levels with 20 trials per level. The decoy sweep inserts an off-target patch into the optical reference under two realizations, an optical cut-and-paste of the anchor's true footprint and a synthetic insertion of the anchor's normalized XRF template, across decoy intensity and decoy count. Scenario definitions, sweep ranges, and matcher settings are documented in Appendix~\ref{app:perturbation_protocols}.

\section{Results and Discussion}
\label{sec:results_discussions}

\subsection{Isolation case study}
\label{sec:isolation_result}

The isolation case study is designed to separate non-geometric per-tile localization from geometry-aware group localization. We first show the outcome, then explain it through a perturbation sweep, and finally check that it survives changing the local similarity criterion.

\begin{figure}[!htbp]
  \centering
  \includegraphics[width=0.98\linewidth]{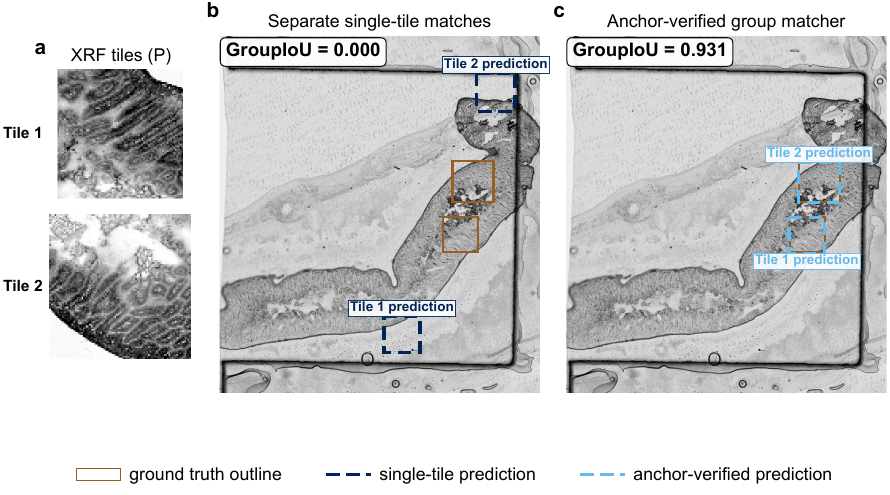}
  \caption{Designed-ambiguity example from the isolation case study. Left: XRF tiles. Middle: independent per-tile localization places both tiles in visually similar but wrong regions (GroupIoU 0.000). Right: anchor-verified group localization correctly places the full group (GroupIoU 0.931).}
  \label{fig:controlled_geometry}
\end{figure}

\begin{figure}[!htbp]
  \centering
  \includegraphics[width=0.98\linewidth]{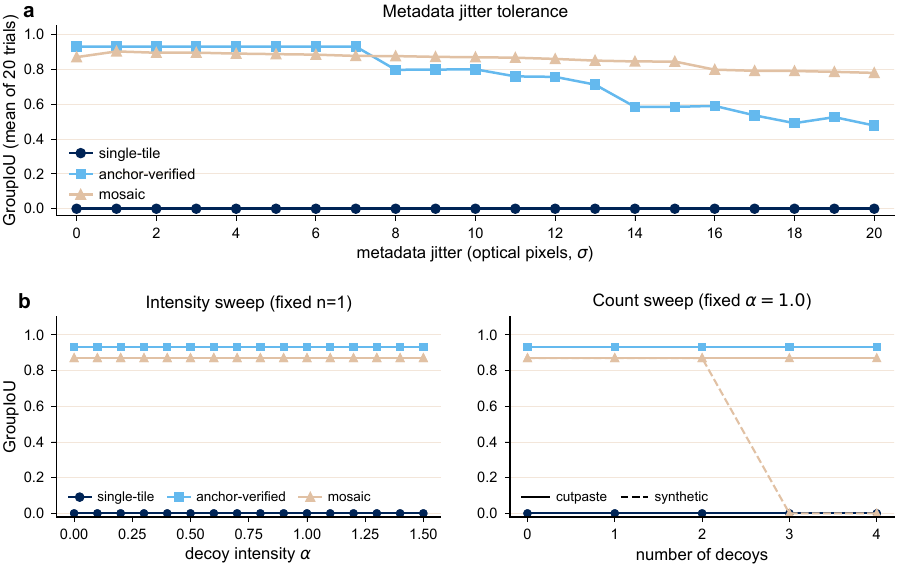}
  \caption{Isolation-case-study perturbation sweeps on Tile~1 and Tile~2 under normalized cross-correlation. (a)~GroupIoU as a function of Gaussian metadata-position jitter applied to the companion tile's recorded scan center, mean of 20 trials per level. Independent per-tile localization is included as an invariant reference because it does not consume acquisition metadata; anchor-verified and mosaic localization show the soft-constraint and hard-constraint responses to metadata error. A same-colored ``x'' indicates that at least one trial at that level returned no-match. (b)~Decoy intensity (left) and decoy count (right) sweeps under two decoy realizations, cut-and-pasted optical patch at the anchor footprint (solid) and synthetic insertion of the anchor's normalized XRF template (dashed). Cut-and-pasted patches do not act as cross-modal decoys, while multiple synthetic insertions collapse the fused-mosaic strategy.}
  \label{fig:controlled_mechanism}
\end{figure}

In a two-tile example (Fig.~\ref{fig:controlled_geometry}) with visually similar local image regions, independent per-tile localization places both tiles in the wrong regions (GroupIoU 0.000). Anchor-verified group localization uses the same image content, but after choosing an anchor candidate it requires the companion tile to appear near the position recorded by the scan metadata. This extra requirement is sufficient to recover the correct placement (GroupIoU 0.931). To investigate why geometry helps, we carried out two perturbation sweeps on this pair (Fig.~\ref{fig:controlled_mechanism}).

The metadata-jitter sweep probes the cost of consuming acquisition metadata when that metadata is unreliable. Here $\sigma_{\mathrm{px}}$ is the per-coordinate standard deviation of the zero-mean Gaussian perturbation applied to the recorded companion-tile center, expressed in equivalent optical pixels. Independent per-tile localization is shown as an invariant non-geometric reference, while the two metadata-consuming strategies change as the metadata they depend on is perturbed. Anchor-verified group localization is flat at 0.931 up to $\sigma_{\mathrm{px}}\approx 7$, transitions over 8--14~px (reaching 0.585 at $\sigma_{\mathrm{px}}{=}14$), and decays further to 0.478 by $\sigma_{\mathrm{px}}{=}20$. The same-colored ``x'' markers indicate levels at which at least one of the 20 trials returned a no-match rather than a high-confidence wrong answer. Mosaic localization degrades more slowly on this real pair, decaying from 0.870 at $\sigma_{\mathrm{px}}{=}0$ to 0.780 at $\sigma_{\mathrm{px}}{=}20$. The two single-scale strategies span a complementary trade-off in which mosaic accepts a globally miscalibrated arrangement while anchor-verified localization rejects companion locations that violate the recorded geometry.

The decoy sweep tests whether off-target patches can attract each strategy away from the true group placement. At the tested intensities and counts, cut-and-pasted optical patches do not act as cross-modal decoys. This result is consistent with the matcher comparing XRF gradient templates against optical gradient patterns rather than raw optical appearance. Only the synthetic-template realization attracts the mosaic matcher strongly enough to collapse the fused-template strategy when multiple decoys are inserted. Anchor-verified localization remains at 0.931 across the decoy sweep because the companion tile must still verify near the anchor-implied position.

\FloatBarrier

A natural concern is that the improvement above depends on the particular local similarity score rather than on acquisition geometry. To test this sensitivity, we re-ran the isolation case study with mutual information (MI) in place of normalized cross-correlation (NCC) while keeping every other part of the search procedure fixed. Anchor-verified group localization is essentially unchanged by the metric swap. In the auto-channel setting, GroupIoU is 0.927 under both NCC and MI for $n{=}2$, and 0.924 under both metrics for $n{=}3$. In the fixed-P setting, GroupIoU is 0.931 under NCC and 0.927 under MI for $n{=}2$, and 0.927 under both metrics for $n{=}3$. These results show that the outcome is insensitive to these two local similarity criteria. Together with the matched independent-versus-group comparison, the metric control supports attributing the improvement to acquisition geometry rather than to a metric-specific effect.
\FloatBarrier

\subsection{Multiscale case study}
\label{sec:multiscale_result}

The multiscale case study tests whether an intermediate coarse-scan bridge image extends the same group-localization idea across spatial scales. We compare four placements per case: anchor-verified localization without a bridge, the two bridge-informed strategies introduced in \S\ref{sec:multiscale_localization}, and the runtime-validity selection between them. We then complement the GroupIoU summary with visual placement of all four cases to show what GroupIoU does and does not capture.

\begin{figure}[!htbp]
  \centering
  \includegraphics[width=0.98\linewidth]{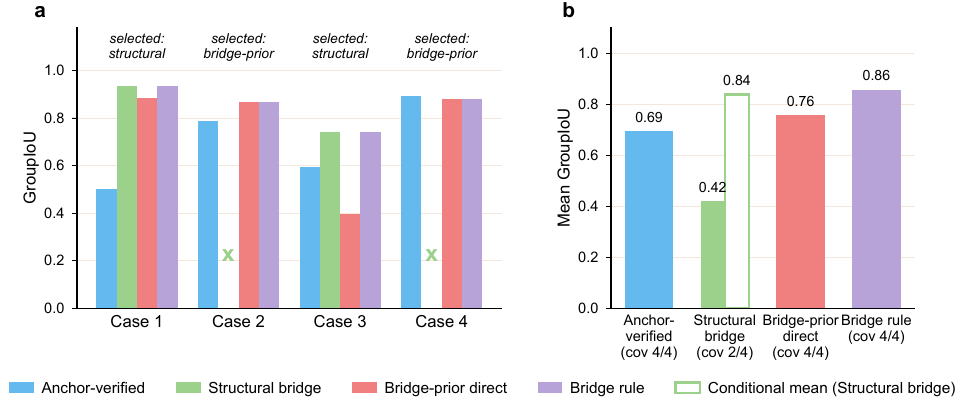}
  \caption{Multiscale case-study summary. (a)~Per-case GroupIoU for the four placements, the italic header above each case names the route selected by the runtime-validity rule, and an ``x'' indicates a no-match return. (b)~Coverage-aware mean GroupIoU across the four cases per route, the filled bar is the failure-aware mean (no-match counted as zero), and a hollow overlay is drawn only for the structural bridge route (coverage 2/4) to show its conditional mean over returned matches. Routes with full 4/4 coverage have failure-aware and conditional means equal by construction and are drawn as a single bar. Numeric value and coverage fraction are annotated above each bar. The full per-case route-level diagnostics are in Table~\ref{tab:appendix_case_diagnostics}.}
  \label{fig:multiscale_summary}
\end{figure}

Anchor-verified localization without a bridge covers all four cases (Fig.~\ref{fig:multiscale_summary}a), with GroupIoU ranging from 0.502 to 0.891 and a mean of 0.694 (Fig.~\ref{fig:multiscale_summary}b). This is a non-trivial baseline (two of the four cases land above 0.78), but it is not uniformly reliable. The two bridge strategies help different cases. Structural-bridge localization returns a valid group placement for Case~1 and Case~3 and is highly accurate when it succeeds (GroupIoU 0.934 and 0.741). Bridge-prior direct localization covers all four cases and is especially useful for Case~2, where the structural route does not return a runtime-valid result. The runtime-validity rule (Eq.~\ref{eq:bridge_rule}) selects the structural route for Case~1 and Case~3 and the bridge-prior route for Case~2 and Case~4, yielding a mean GroupIoU of 0.856 (Fig.~\ref{fig:multiscale_summary}b, ``Bridge rule''). Across the four cases, this is a clear improvement over the no-bridge baseline, but the per-case panel also keeps the boundary visible: in Case~4, anchor-verified localization without a bridge is slightly better than the selected bridge-informed result (0.891 versus 0.879). The reported gain comes from complementary bridge roles across the four cases, not from a universally dominant route.

\begin{figure}[!b]
  \centering
  \includegraphics[width=0.78\linewidth]{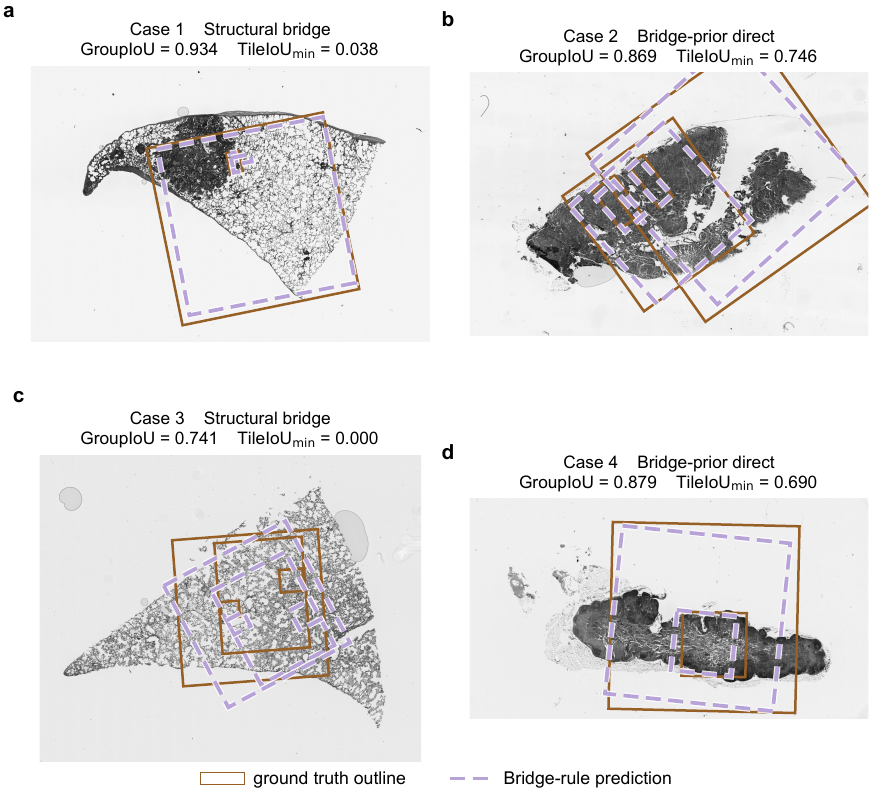}
  \caption{Selected placements for the four cases in the multiscale case study. Each panel shows the optical image with ground-truth footprints (solid outlines) and Bridge-rule predicted footprints (light-purple dashed outlines). Panel titles name the route selected by the Bridge rule: Cases 1 and 3 use structural-bridge localization, while Cases 2 and 4 use bridge-prior direct localization.}
  \label{fig:case_progression}
\end{figure}

For completeness, we tested standard keypoint pipelines as a possible alternative. SIFT returns GroupIoU 0.000 on all four cases, and ORB reaches at most 0.019. This is expected because elemental XRF maps and optical images emphasize different physical structures, so the strongest keypoints in one modality rarely have stable counterparts in the other. Off-the-shelf descriptor localization is therefore not a drop-in replacement for the template-and-geometry combination evaluated here.

Fig.~\ref{fig:case_progression} shows the selected placements overlaid on the optical image. Cases 1, 2, and 4 are relatively accurately placed visually: the light-purple prediction outlines closely follow the solid ground-truth outlines. Case 3 (Fig.~\ref{fig:case_progression}c) is the clearest failure in the study. Its bridge-rule GroupIoU is moderate (0.741), but the per-tile overlap collapses (TileIoU$_\text{min}=0.000$, appendix Table~\ref{tab:appendix_case_diagnostics}) because the predicted group is rotated relative to ground truth. The same pattern in a milder form appears in Case 2 and 4 as well. The shared global similarity transform can absorb a small in-plane rotation while still covering most of the correct FOV at the union level. GroupIoU does not penalize this, but per-tile overlap does. These cases motivate a two-level reporting practice: GroupIoU for FOV assessment, and tile-level diagnostics whenever downstream interpretation depends on individual elemental maps. A closer diagnostic of Case 3 (Appendix~\ref{app:appendix_case3_failure}) indicates that the target group remains comparatively well localized in the bridge frame, but a bridge-to-optical ambiguity transfers a rotation into the final optical-frame placement. Limited nonrigid group adjustment or learned cross-modal features could plausibly mitigate this failure mode, but both are outside the scope of this study.

\section{Conclusions}
\label{sec:conclusion}

This work argues for a change in perspective: the natural unit of XRF-to-optical localization is the tile group, not the individual tile. When the instrument has already recorded how XRF tiles are arranged relative to each other, that arrangement is localization evidence as much as the image content itself. Treating the tiles as a structured group turns acquisition metadata into a direct check on candidate placements, because companion tiles must appear near the positions recorded by the scan metadata, and it gives a meaningful way to ask whether the whole XRF FOV has been placed correctly.

The isolation case study shows that this check resolves ambiguous local matches that the non-geometric independent per-tile reference cannot distinguish. The NCC-versus-MI control shows that this outcome is not specific to either local similarity metric, while the matched comparison supports attributing the improvement to acquisition geometry. The same study traces the operating envelope. The anchor-verified group score remains flat up to about seven optical pixels of metadata-position error, degrades beyond that, and begins to return no-match outcomes. Mosaic localization degrades more slowly under jitter on the real pair but is fragile when tiles do not share a common scan-grid size. In the controlled tests, the two single-scale strategies therefore span complementary trade-offs: anchor-verified group localization favors accurate per-tile placement, while mosaic favors group-level FOV placement when the fused template is reliable. The multiscale case study extends the same idea across spatial scales, with an intermediate coarse XRF scan and two complementary routes selected at runtime by a validity check. Reporting GroupIoU together with tile-level diagnostics surfaces failure modes the union score alone would hide.

The present workflow has three practical implications for acquisition. Imaging the same physical specimen by optical and XRF microscopy avoids the section-to-section ambiguity present in the multiscale dataset. A coarse, large-step XRF survey scan provides the bridge between the optical image and finer XRF scans. At least two fine-scale XRF tiles are required for their relative acquisition geometry to constrain localization beyond an independent single-tile match.

Several limitations remain. The shared global similarity transform excludes nonrigid deformation. The case studies use grayscale intensity and gradient-magnitude representations. Learned cross-modal representations could improve local image similarity and could be combined with the geometry-aware framework in the future. Longer bridge chains or multiple bridges per group are a natural extension for future work.

\newpage
\appendix

\section{Mathematical Formulation}
\label{app:math_formulation}

\begin{table}[!htbp]
  \centering
  \caption{Notation used in the group-localization formulation.}
  \label{tab:method_notation}
  \small
  \setlength{\tabcolsep}{5pt}
  \begin{tabular}{p{0.22\linewidth} p{0.70\linewidth}}
    \toprule
    Symbol & Meaning \\
    \midrule
    $I$ & Optical reference image. \\
    $J$ & Intermediate XRF bridge image, when available. \\
    $\mathcal{T}=\{T_1,\dots,T_n\}$ & XRF tile group acquired from one specimen, with each $T_i$ a 2D template. \\
    $x_i$ & Physical scan center of tile $T_i$ recorded by the instrument. \\
    $a$ & Index of the anchor tile (the tile whose appearance proposes the placement). \\
    $\theta$ & Candidate group placement: anchor center, scale, orientation, optional flip. \\
    $P_\theta$ & Offset map induced by $\theta$ that converts physical scan offsets into optical-pixel offsets. \\
    $\widehat{c}_i(\theta)$, $\widehat{B}_i(\theta)$ & Predicted center and predicted footprint of tile $T_i$ under $\theta$. \\
    $B_i^\star(\theta)$ & Locally verified footprint of tile $T_i$ near its predicted footprint. \\
    $\phi_i(B,I)$ & Local image similarity between tile $T_i$ and the optical crop in footprint $B$. \\
    $M(\theta)$ & Tiles that pass local verification and runtime validity checks. \\
    $S_{\mathrm{grp}}(\theta)$ & Whole-group score used to rank candidate placements. \\
    \bottomrule
  \end{tabular}
\end{table}

\paragraph{Single-tile baseline.}
Single-tile localization selects the placement that maximizes the local image similarity for one anchor template,
\begin{equation}
\widehat{\theta}_{\mathrm{single}}
=\operatorname*{arg\,max}_{\theta\in\Theta_a}\phi_a(B_a(\theta),I),
\label{eq:single_tile_objective}
\end{equation}
where $\phi_a$ scores the anchor template against the optical crop inside the anchor footprint $B_a(\theta)$.

\paragraph{Local tile verification and center residual.}
Given a candidate placement $\theta$, each companion tile is verified inside a local search window around its predicted footprint:
\begin{equation}
B_i^\star(\theta)=
\operatorname*{arg\,max}_{B\in\mathcal{N}(\widehat{B}_i(\theta))}
\left[\phi_i(B,I)-\beta\,d_c\left(B,\widehat{B}_i(\theta)\right)\right],
\label{eq:tile_verification}
\end{equation}
where $\mathcal{N}(\widehat{B}_i(\theta))$ is the local window, $d_c$ is the normalized distance between footprint centers, and $\beta$ discourages large excursions from the recorded prediction. The corresponding center residual is
\begin{equation}
r_i(\theta)=d_c\left(B_i^\star(\theta),\widehat{B}_i(\theta)\right).
\label{eq:center_residual}
\end{equation}

\paragraph{Whole-group score.}
Let $M(\theta)$ denote the tiles that pass local verification and runtime validity checks. The whole-group score is
\begin{equation}
S_{\mathrm{grp}}(\theta)=
\frac{1}{W_{\mathcal{T}}}\sum_{i\in M(\theta)}w_i\phi_i(B_i^\star(\theta),I)
-\lambda_r\frac{1}{W_{M}}\sum_{i\in M(\theta)\setminus\{a\}}w_i r_i(\theta)
-\lambda_o O(\theta),
\label{eq:group_score}
\end{equation}
where $W_{\mathcal{T}}=\sum_i w_i$ and $W_M=\sum_{i\in M(\theta)\setminus\{a\}}w_i$ are the expected and matched total tile weights, $w_i$ are uniform or scaled by physical tile support, and $O(\theta)$ penalizes disagreement between predicted and verified overlap patterns. We note that equation~(\ref{eq:group_score}) is evaluated only for candidates with at least two matched tiles, including at least one verified companion tile, so $W_M>0$. The asymmetric normalization is intentional: dividing the similarity sum by the expected weight $W_\mathcal{T}$ makes partial matches accumulate less score than full matches, while the residual penalty is averaged only over tiles where a residual was actually measured. Candidates that fail minimum tile-count, matched-weight, or visibility checks are returned as a no-match.

\paragraph{Single-scale group strategies.}
Mosaic localization first fuses the tiles into a single template in group coordinates,
\begin{equation}
T_{\mathcal{T}}(u)=
\frac{\sum_i m_i(u)\,T_i(G_i^{-1}u)}
{\sum_i m_i(u)+\epsilon},
\label{eq:mosaic_template}
\end{equation}
using per-tile transforms $G_i$ and valid-pixel masks $m_i$, and then matches the mosaic as a single template,
\begin{equation}
\widehat{\theta}_{\mathrm{mosaic}}
=\operatorname*{arg\,max}_{\theta\in\Theta_{\mathcal{T}}}
\phi_{\mathcal{T}}(B_{\mathcal{T}}(\theta),I).
\label{eq:mosaic_objective}
\end{equation}
Anchor-verified group localization keeps tiles separate during scoring. For a fixed anchor $a$ with anchor-candidate set $\mathcal{C}_a$,
\begin{equation}
\widehat{\theta}_{\mathrm{group},a}
=\operatorname*{arg\,max}_{\theta\in\mathcal{C}_a}S_{\mathrm{grp}}(\theta),
\label{eq:anchor_verify_objective}
\end{equation}

\paragraph{Multiscale group strategies.}
Structural-bridge localization first matches the bridge image $J$ to the optical frame and then matches the target group inside the bridge, composing the two placements:
\begin{equation}
\widehat{\psi}=\operatorname*{arg\,max}_{\psi} \phi_B(B_B(\psi),I),
\qquad
\widehat{\omega}=\operatorname*{arg\,max}_{\omega} S_{\mathrm{grp}}(\omega,J,\mathcal{T}),
\qquad
\widehat{\theta}_{\mathrm{struct}}=\widehat{\psi}\circ\widehat{\omega}.
\label{eq:structural_bridge}
\end{equation}
Here $\psi$ ranges over bridge-placement candidates, $B_B(\psi)$ is the bridge footprint in optical coordinates under $\psi$, and $\phi_B$ is the local similarity between the bridge template and the optical crop in $B_B(\psi)$. The target-group search then ranges $\omega$ over target placements inside the bridge image, scored by the whole-group score $S_{\mathrm{grp}}$ with $J$ standing in for the optical reference. Bridge-prior direct localization uses the bridge as a spatial prior: if $\theta_0(\widehat{\psi})$ is the target placement implied by the bridge-to-optical estimate and the bridge-to-target metadata, then
\begin{equation}
\widehat{\theta}_{\mathrm{prior}}
=\operatorname*{arg\,max}_{\theta\in\mathcal{N}(\theta_0)}
\left[
S_{\mathrm{grp}}(\theta,I,\mathcal{T})
-\lambda_b d_{\Theta}(\theta,\theta_0)
\right],
\label{eq:bridge_prior_direct}
\end{equation}
with $\mathcal{N}(\theta_0)$ a bridge-derived search neighborhood and $\lambda_b d_{\Theta}$ an optional prior penalty.

\section{Algorithm Details}
\label{app:algorithm_details}

\begin{table}[!htbp]
  \centering
  \caption{Geometry-aware grouped localization. The procedure first proposes anchor placements, then asks whether the companion tiles can be verified near the locations recorded by acquisition geometry.}
  \label{tab:algorithm_group_localization}
  \small
  \setlength{\tabcolsep}{5pt}
  \begin{tabular}{>{\raggedright\arraybackslash}p{0.08\linewidth} >{\raggedright\arraybackslash}p{0.82\linewidth}}
    \toprule
    Step & Operation \\
    \midrule
    1 & Prepare the optical reference image and the XRF tile templates. Read each tile center and pixel size from acquisition metadata. \\
    2 & Select an anchor tile or an eligible set of anchor tiles. Generate anchor candidates by multiscale template localization and non-maximum suppression. \\
    3 & For each anchor candidate, use Eq.~\ref{eq:center_prediction} to predict the footprint of every companion tile from the recorded group geometry. \\
    4 & Search a local window around each predicted companion footprint. Keep the best local match and record its image score and center residual. \\
    5 & Form the matched set $M(\theta)$ from tiles that pass local score, visibility, and support checks. Reject candidates that fail minimum tile-count or match-weight requirements. \\
    6 & Score each valid candidate with Eq.~\ref{eq:group_score}. Select the highest-scoring valid group placement, or report no match if no candidate passes the runtime checks. \\
    \bottomrule
  \end{tabular}
\end{table}

\begin{table}[!htbp]
  \centering
  \caption{Bridge-informed route selection. The structural and bridge-prior routes are run as separate localization paths before the runtime-validity rule selects one returned placement.}
  \label{tab:algorithm_bridge_rule}
  \small
  \setlength{\tabcolsep}{5pt}
  \begin{tabular}{>{\raggedright\arraybackslash}p{0.08\linewidth} >{\raggedright\arraybackslash}p{0.82\linewidth}}
    \toprule
    Step & Operation \\
    \midrule
    1 & Match the bridge image to the optical image to estimate the bridge-to-optical placement. \\
    2 & Run the structural bridge route by localizing the target XRF group inside the bridge image and composing the target-to-bridge and bridge-to-optical placements. \\
    3 & Independently run the bridge-prior direct route by using the bridge placement to define an optical-frame search neighborhood, then localizing the target group directly in the optical frame. \\
    4 & If the structural route returns a runtime-valid group result, select it. Otherwise select the bridge-prior direct result. \\
    5 & Use ground-truth overlap only after selection, for evaluation and reporting. \\
    \bottomrule
  \end{tabular}
\end{table}

The independent per-tile reference runs the same local tile scorer separately for each tile and reports the union of the returned footprints. Mosaic localization first fuses the tile group into one template using acquisition geometry, then runs ordinary template localization.

\section{Multiscale Experiment Settings and Per-Case Diagnostics}
\label{app:case_level_diagnostics}

This appendix documents the fixed evaluation settings used for all multiscale bridge-informed results and the per-case route-level diagnostics behind the coverage-aware summary in the main text. Table~\ref{tab:released_settings} lists the settings that determine how candidates are generated, verified, and rejected. 

\begin{table}[!htbp]
  \centering
  \caption{Settings and runtime validity checks used for the multiscale bridge-informed results.}
  \label{tab:released_settings}
  \small
  \setlength{\tabcolsep}{4pt}
  \begin{tabular}{>{\raggedright\arraybackslash}p{0.28\linewidth} >{\raggedright\arraybackslash}p{0.22\linewidth} >{\raggedright\arraybackslash}p{0.40\linewidth}}
    \toprule
    Setting or check & Reported value & Role \\
    \midrule
    XRF normalization & 2--98\% percentile range & Robust grayscale normalization before localization. \\
    Isotropic XRF resampling & enabled, geomean step & Converts anisotropic scan pixels to a common isotropic grid. \\
    Stage-1 bridge-to-optical feature and metric & gradient magnitude, NCC & Localizes the bridge image in the optical frame. \\
    Stage-1 angle search & center 0$^\circ$, range 60$^\circ$, step 6$^\circ$ & Searches bridge orientation around the nominal direction. \\
    Stage-2 target-to-bridge feature and metric & raw XRF, NCC & Matches the target group inside the bridge image for the structural route. \\
    Stage-2 scale search & one scale, 5\% span & Uses metadata-centered scale for target-to-bridge localization. \\
    Stage-2 angle search & range 45$^\circ$, step 5$^\circ$ & Searches target orientation inside the bridge image. \\
    Final direct feature and metric & gradient magnitude, MI & Scores direct target-group placement in the optical frame for the bridge-prior route. \\
    Bridge-prior seed weight & 0.25 & Weights the bridge-derived proposal in the direct route. \\
    Local center penalty & 0.25 & Penalizes companion-tile displacement from recorded locations during local verification. \\
    Group center penalty & 0.20 & Penalizes group-level center residuals in the final score. \\
    Minimum matched tiles & 2 & Rejects group placements with too little tile support. \\
    Minimum matched fraction and weight fraction & 0.6 and 0.6 & Requires sufficient unweighted and weighted tile support. \\
    Minimum per-tile visible area fraction & 0.05 & Counts a predicted tile as visible when at least 5\% of its footprint lies within the reference-image bounds. \\
    Minimum visible fraction and visible weight fraction & 0.6 and 0.6 & Rejects placements with insufficient visible predicted group support. \\
    Bridge-rule validity & structural stage-2 group result returned & Selects the structural route only when runtime checks return a valid group result. \\
    \bottomrule
  \end{tabular}
\end{table}

Table~\ref{tab:appendix_case_diagnostics} gives the route-level results behind the coverage-aware summary in the main text. The selected column records the route used by the bridge rule. The status column reports runtime outcome before evaluation. GroupIoU measures FOV placement, while TileIoU columns show whether individual tile footprints were also localized well.

\begin{table}[!htbp]
  \centering
  \caption{Route-level diagnostics for the four multiscale cases. No-match entries did not return a final group prediction.}
  \label{tab:appendix_case_diagnostics}
  \small
  \setlength{\tabcolsep}{3.5pt}
  \begin{tabular}{l l c c c c c c}
    \toprule
    Case & Route & Status & Selected & GroupIoU & TileIoU$_\text{min}$ & TileIoU$_\text{mean}$ & TileIoU$_\text{wmean}$ \\
    \midrule
    Case 1 & Anchor-verified & ok & no & 0.502 & 0.000 & 0.167 & 0.498 \\
    Case 1 & Structural bridge & ok & yes & 0.934 & 0.038 & 0.409 & 0.927 \\
    Case 1 & Bridge-prior direct & ok & no & 0.883 & 0.000 & 0.294 & 0.876 \\
    Case 2 & Anchor-verified & ok & no & 0.788 & 0.048 & 0.444 & 0.711 \\
    Case 2 & Structural bridge & no match & no & -- & -- & -- & -- \\
    Case 2 & Bridge-prior direct & ok & yes & 0.869 & 0.746 & 0.810 & 0.864 \\
    Case 3 & Anchor-verified & low confidence & no & 0.594 & 0.000 & 0.297 & 0.578 \\
    Case 3 & Structural bridge & ok & yes & 0.741 & 0.000 & 0.327 & 0.660 \\
    Case 3 & Bridge-prior direct & partial & no & 0.395 & 0.000 & 0.169 & 0.348 \\
    Case 4 & Anchor-verified & ok & no & 0.891 & 0.833 & 0.862 & 0.884 \\
    Case 4 & Structural bridge & no match & no & -- & -- & -- & -- \\
    Case 4 & Bridge-prior direct & ok & yes & 0.879 & 0.690 & 0.785 & 0.858 \\
    \bottomrule
  \end{tabular}
\end{table}

\section{Perturbation Protocol Details}
\label{app:perturbation_protocols}

This appendix documents the sweep ranges and matcher settings for the two perturbation protocols introduced in Section~\ref{sec:perturbation_protocols}. Both protocols share a common matcher configuration. Single-tile localization is run independently per tile, and the union of the two predictions is reported. Anchor-verified group localization uses a local-verification window of 24 optical pixels and a group-center penalty weight of 0.5. The local-verification window corresponds to about 12\% of a tile's ground-truth footprint width and is chosen so that the geometry constraint is discriminating without being unrealistically restrictive. Mosaic localization is run with its default configuration.

\paragraph{Metadata-position jitter sweep.}
For each level $\sigma_\mathrm{px}\in\{0,1,\dots,20\}$ optical pixels and each of 20 independent trials, we draw a two-dimensional Gaussian perturbation $\Delta x_i\sim\mathcal{N}(0,\sigma_\mathrm{um}^2)$ on every non-anchor tile's recorded scan center, with $\sigma_\mathrm{um}=\sigma_\mathrm{px}\cdot(\mathrm{step}_\mathrm{um}/\mathrm{scale})$ converting an equivalent optical-pixel jitter into the metadata units actually perturbed. The optical reference and tile content are unchanged across trials.

\paragraph{Decoy sweep.}
A decoy is a square patch the size of the anchor's ground-truth footprint, inserted into the optical reference at one or more pre-chosen off-target locations away from either ground-truth footprint. The cutpaste variant copies the optical-image content at the anchor's true footprint and pastes it at the off-target locations via pixel-wise maximum with intensity scaling $\alpha$. The synthetic variant resamples the anchor's normalized XRF template to the same patch size and inserts it in the same way. The intensity axis sweeps $\alpha\in\{0.0,0.1,\dots,1.5\}$ with one decoy at the first off-target location, and the count axis sweeps the number of decoys in $\{0,1,2,3,4\}$ at $\alpha{=}1.0$. The two-variant design separates effects driven by the decoy realization from effects driven by the decoy load.

\section{Stress test on a 3-Tile Combination}
\label{app:mechanism_appendix_3tile}

\begin{figure}[!htbp]
  \centering
  \includegraphics[width=0.80\linewidth]{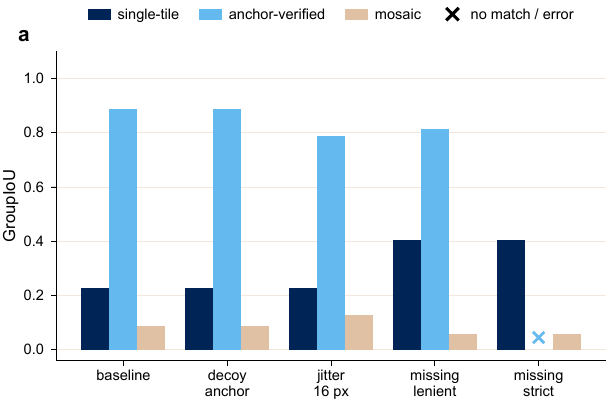}
  \caption{Stress test on the real 3-tile combination. The scenarios include baseline, synthetic decoy, metadata jitter, and two missing-tile settings in which one tile's optical region is zeroed out and the group matcher is run under permissive (lenient) or strict matched-fraction thresholds. A same-colored ``x'' marks an error or no-match return.}
  \label{fig:mechanism_appendix_3tile}
\end{figure}

Fig.~\ref{fig:mechanism_appendix_3tile} runs an expanded stress set on a 3-tile combination. It includes the unperturbed baseline, a synthetic decoy, a metadata-position jitter setting at $\sigma_{\mathrm{px}}{=}16$, and two missing-tile scenarios in which the optical region under one tile's ground-truth footprint is zeroed out. In the lenient missing-tile setting the matched-fraction threshold and per-tile score floor are relaxed, so the group strategy accepts a partial placement supported by the remaining tiles. In the strict setting the same thresholds are tightened, so the group strategy returns a no-match instead of a partial-support placement. The anchor-verified group strategy outperforms the independent per-tile reference and mosaic at baseline, retains baseline-level GroupIoU under the decoy perturbation, degrades under jitter, and explicitly returns a no-match under the strict missing-tile policy. Mosaic localization degrades sharply at baseline on this 3-tile combination because the three tiles do not share a common scan-grid size and the fused template is not a faithful composite, a known fragility of the hard-constraint variant when tile shapes differ.

\section{Case 3 Failure Analysis}
\label{app:appendix_case3_failure}

\begin{figure}[!htbp]
  \centering
  \includegraphics[width=0.98\linewidth]{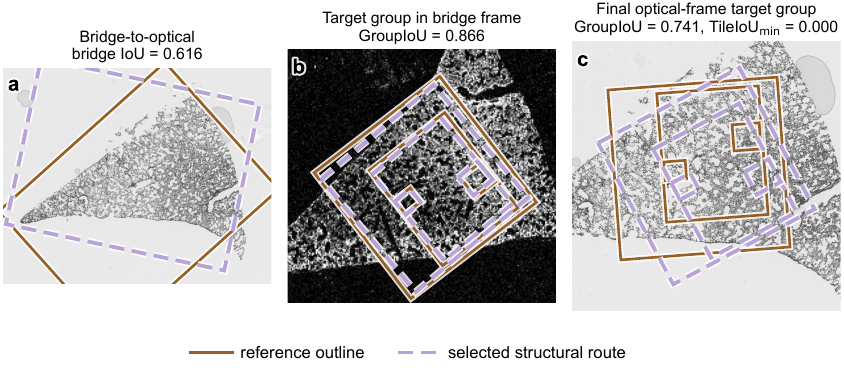}
  \caption{Case~3 failure diagnostic. (a)~The selected structural route first localizes the bridge image in the optical frame. The bridge footprint overlaps the correct tissue region but is rotated relative to the reference outline. (b)~In the bridge frame, the target group placement is substantially better aligned. (c)~Composing the target-to-bridge result with the bridge-to-optical placement transfers the rotation into the optical frame. The resulting union overlap remains moderate, but one tile has zero overlap with its ground-truth footprint.}
  \label{fig:case3_failure_diagnostic}
\end{figure}

Fig.~\ref{fig:case3_failure_diagnostic} separates the selected structural route for Case~3 into the two transformations that are composed in the final placement. The bridge-to-optical step returns a plausible but rotated bridge footprint, with bridge IoU 0.616 (Fig.~\ref{fig:case3_failure_diagnostic}a). In contrast, the target group remains comparatively well aligned inside the bridge frame, with GroupIoU 0.866 (Fig.~\ref{fig:case3_failure_diagnostic}b). After composition into the optical frame, the final selected placement retains moderate whole-FOV overlap (GroupIoU 0.741), but the per-tile minimum overlap collapses to 0.000 because the tile group is rotated relative to the ground truth (Fig.~\ref{fig:case3_failure_diagnostic}c).

This diagnostic suggests that a major Case~3 error enters through the bridge-to-optical ambiguity rather than through a loss of target-group geometry inside the bridge frame. The case is therefore a limitation of the current global-similarity and image-similarity formulation. A bridge placement can be close enough to cover much of the correct tissue FOV while still transferring a rotation that matters for individual high-resolution XRF tiles. This is why the main text reports GroupIoU together with tile-level diagnostics: the union score captures FOV placement, while TileIoU exposes whether individual elemental maps are trustworthy for downstream interpretation.

\section*{Acknowledgments}
This research used resources of the Advanced Photon Source, a U.S.~Department of Energy (DOE) Office of Science user facility at Argonne National Laboratory.
The authors thank the staff of APS beamlines 8-BM-B and 2-ID-E, particularly Evan Maxey and Olga Antipova, for support with the XRF measurements.

\section*{Funding}
This research was supported by the U.S. DOE Office of Science, Basic Energy Sciences, under Contract No.~DE-AC02-06CH11357. The authors acknowledge funding from the DOE Office of Science, Basic Energy Sciences, under award 0000283133. Additional support came from the National Institutes of Health, National Institute of Allergy and Infectious Diseases, under award 1P01AI165380-01.

\section*{Conflicts of Interest}
The authors declare no conflicts of interest.

\section*{Data and Code Availability}
The data and code supporting the findings of this study will be released upon publication.

\bibliographystyle{unsrt}
\bibliography{references}

\end{document}